\documentclass[letterpaper, 10 pt, conference]{ieeeconf}  
\IEEEoverridecommandlockouts 
\usepackage[noadjust]{cite} 

\usepackage{amsmath,amsfonts,amssymb} 
\usepackage{graphicx}   
\usepackage{array}
\usepackage{tikz}
\newcommand{\circmark}[1]{%
  \tikz[baseline=(char.base)]{%
    \node[shape=circle, fill=black, inner sep=1.5pt, text=white, font=\small\bfseries] (char) {#1};%
  }%
}

\newcommand{\circlethree}{\circmark{3}}
\newcommand{\circlefour}{\circmark{4}}
\usepackage{booktabs}   
\usepackage{tcolorbox}
\usepackage{xcolor}
\let\labelindent\relax
\usepackage{enumitem}   
\usepackage{mathtools} 
\usepackage{siunitx} 
\usepackage{xspace} 
\usepackage{textcomp} 
\usepackage{url} 
\usepackage{multirow}
\usepackage{graphicx} 
\usepackage{booktabs} 
\usepackage{array} 
\usepackage{stfloats} 
\usepackage[caption=false,font=normalsize,labelfont=sf,textfont=sf]{subfig} 

\usepackage{algorithm} 
\usepackage{algorithmic} 

\title{\Large\bf
RESCORE: LLM-Driven Simulation Recovery in Control Systems Research Papers
}
\author{Vineet Bhat*, Shiqing Wei*, Ali Umut Kaypak, Prashanth Krishnamurthy, \\ Ramesh Karri, and Farshad Khorrami
\thanks{The authors are with ECE Dept, Tandon School of Engineering, New York University, Brooklyn, NY 11201, USA. E-mail: \{vrb9107, shiqing.wei, ak10531, prashanth.krishnamurthy, rkarri, khorrami\}@nyu.edu.}
}

\begin{document}

\maketitle
\thispagestyle{empty}
\pagestyle{empty}

\begin{abstract}
Reconstructing numerical simulations from control systems research papers is often hindered by underspecified parameters and ambiguous implementation details. We define the task of Paper-to-Simulation Recoverability: the ability of an automated system to generate executable code that faithfully reproduces a paper’s results. We curate a benchmark of 500 papers from the IEEE Conference on Decision and Control (CDC) and propose RESCORE: a three-component LLM-agentic framework—Analyzer, Coder, and Verifier. RESCORE uses iterative execution feedback and visual comparison to improve reconstruction fidelity. Our method successfully recovers task-coherent simulations for 40.7\% of benchmark instances, outperforming single-pass generation. Notably, the RESCORE automated pipeline achieves an estimated 10$\times$ speedup over manual human replication, drastically cutting the time and effort required to verify published control methodologies. We will release our benchmark and agents to foster community progress in automated research replication.
\end{abstract}

\section{Introduction}
Reproducibility is a foundational requirement of open science, yet it remains an acute and largely unsolved challenge in various areas of engineering including control systems.  Legacy papers in the science and engineering domains often suffer from missing or obsolete code, proprietary tool chains, discontinued evaluation platforms, and underspecified experimental details. These gaps hinder independent validation and impede cumulative scientific progress.


Advances in multimodal foundation models have made automated paper-to-simulation recovery increasingly feasible. Literature has shown the efficacy of large language models (LLMs) and vision-language models (VLMs) in  PDF extraction \cite{shen2022vila, clark2016pdffigures, lopez2010humb}, math-aware optical character recognition \cite{deng2017image, blecher2023nougat}, software synthesis~\cite{austin2021program} and research~\cite{tian_scicode}. Emergence of ``paper-to-code'' benchmarks, such as PaperBench \cite{starace2025paperbench} and Paper2Code \cite{seo2025paper2code}, has shown promise in automatic replication of machine learning (ML) research.

In control systems papers, simulations present a distinct, unexplored challenge: success is not defined by passing static unit tests or syntactic correctness, but by recovering a dynamical system whose behavior agrees with that in the paper. Simulation plots serve as a primary medium to communicate system behavior, controller performance, and comparative results. However, evaluating the computational validity of these results is hindered because publishing executable code is not the norm in this community~\cite{how2018control}. 

A  body of work on scientific computing distinguishes  \emph{computational reproducibility} (assumes access to code, data, and computational environment) \cite{national2019reproducibility} and the pragmatic reconstruction of a method from the paper. We introduce \emph{Paper-to-Simulation Recoverability}: an operational proxy measuring if a published paper has enough information to reconstruct an executable simulation that agrees with plots.

To tackle this challenge, one needs to go beyond the single-pass code generation baselines \cite{austin2021program, chen2021evaluating, hendrycks2021measuring, li2022competition} used in early program synthesis literature. A central lesson from software engineering and coding-agent benchmarks \cite{jimenez2024swe, yang2024swe, liu2024repobench, du2024evaluating, badertdinov2025swe} is that complex, multi-component tasks require iterative workflows. Drawing on  frameworks for debugging and execution-feedback \cite{chen2024teaching, gehring2025rlef}, we propose RESCORE (\textbf{Re}constructing \textbf{S}imulations from \textbf{Co}ntrol \textbf{Re}search) as a closed-loop agentic architecture with three components: an Analyzer, a Coder, and a Refiner. Instead of relying on a zero-shot generation, RESCORE leverages terminal execution traces and visual plot-level discrepancies to diagnose and repair the reconstructed code, mirroring ``reason-act-observe'' loops effective in agentic systems \cite{yao2023react, gouc2024ritic,bhat_brainbodyllm}.



Robust evaluation of such systems requires thwarting shortcuts \cite{yu2025utboost}, such as an agent plotting a hard-coded curve to mimic a figure without implementing the underlying dynamics or control design. We establish a multi-axis evaluation protocol that combines execution checks, visual agreement metrics, and  LLM-as-a-judge audits to assess specification alignment and equation coverage. Beyond achieving benchmark scores, RESCORE explores a methodology for automated simulation recovery. While manual reconstruction of simulations uses days of effort from researchers, RESCORE offers a huge speed-up. Furthermore, as LLMs evolve, RESCORE can extend to massive technical document repositories, such as comprehensive power systems literature, paving the way for accelerated research validation. 

The contributions of this work are threefold:
\begin{enumerate}
    \item \emph{Paper-to-Simulation Recoverability} as a new benchmark task for control, measuring whether an executable and faithful simulation can be reconstructed from the paper.
    \item A benchmark set and evaluation framework, including select criteria, structured annotations, and multi-axis metrics for execution validity, figure reconstruction quality, equation coverage, and specification alignment.
    \item A feedback-guided LLM-agentic framework that analyzes a paper, generates code, and iteratively fine-tunes it using visual guidance. Closed-loop repair against execution traces and figure-level discrepancies are critical to improve recoverability in control simulations.
\end{enumerate}

\section{RESCORE Methodology}

\begin{figure*}[thpb]
      \centering
      \includegraphics[width=\linewidth]{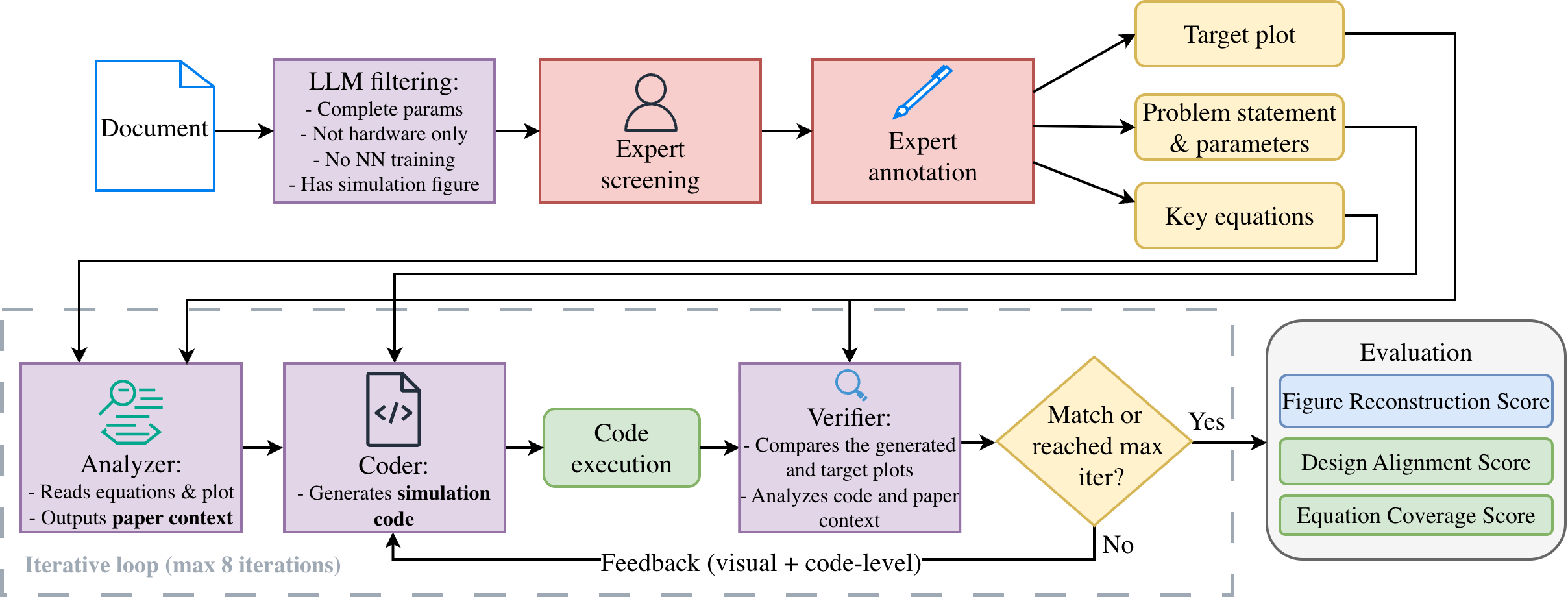}
      \caption{RESCORE framework automates code recovery from control system papers. After filtering, expert screening, and annotation, Analyzer, Coder, and Verifier LLM agents operate in a closed loop to generate, execute, and refine simulation code using feedback,
      followed by evaluation.}
      \label{fig:framework_arch}
   \end{figure*}


Reconstructing executable simulations from control papers requires synthesizing key equations, extracting scattered parameter values, and generating code that reproduces reported behavior. While LLMs possess strong code-generation capabilities, they require grounding in a paper's mathematical and technical context. We propose RESCORE, a multi-agent framework for equation understanding, code generation, and visual feedback, where simulation plots serve as iterative debug signals for 
refinement. Figure~\ref{fig:framework_arch} illustrates the methodology. The agents employ GPT-5.2~\cite{openai2025gpt52} as the core LLM.

\subsection{Paper Selection Criteria}
\label{subsec:paper-selection-criteria}

We construct a database of 100 most cited papers from last five years of IEEE CDC (2021–2025). A paper is \emph{code-recoverable} if four criteria are met: (i) parameter values are complete; (ii) has primary simulations, excluding papers that rely on hardware experiments, third-party simulations, or proprietary data; (iii) does not rely on training neural networks; (iv) includes a clear target simulation figure. These criteria ensure that the  papers are hardware-independent and have self-contained descriptions for code generation. We use Qwen 3 \cite{yang2025qwen3} LLM, to filter the database against these four criteria. A domain expert then reviews the filtered list for a second screening. Finally, 221-out-of-500 papers were marked as code-recoverable and advanced to the generation phase. Analysis on  paper selection is in Section.~\ref{sec:results}.

\subsection{Expert Annotation and Problem Design}
Filtered papers are reviewed by a domain expert, who identifies simulation plots, problem formulations, key equations, and parameter values as context for code recovery via a semi-automated process. This grounds the LLM in the paper's mathematical and technical scope. The step requires $\sim$20 minutes per paper and mirrors the practical use case of a coding assistant following a developer's specifications.




\subsection{Automated Code Recovery}
RESCORE has a closed-loop structure with three agents:

\paragraph{Analyzer} 
receives screenshots of the expert-highlighted equations and the target verification plot, alongside the problem statement and parameter values. It transcribes the equations into plain text and uses the surrounding context to describe the structure of the equation, its parameters, and intended function. It also analyzes the verification plot to describe the system's expected behavior (qualitative,  quantitative). These outputs form the \emph{paper context}. Prompt for the analyzer agent is provided in Figure~\ref{fig:prompt-hint}.

\begin{figure}[t]
  \centering
  \small
  \begin{tcolorbox}[title=Agent 1: Analyzer --- Equation \& Plot Analysis]
    \textbf{System (Equation):}
    You are an expert control systems engineer analysing a research paper annotated with red bounding boxes around equations critical for implementation. Transcribe every equation exactly. Define every variable and symbol.
    Provide chain-of-thought reasoning about relevance.\\

    \textbf{User (Equation):}
    Analyzing page \{page\_num\} of a given research paper.
    Red bounding boxes mark key equations for: \{problem\_statement\}.
    Known parameters: \{params\}, \{init\_conditions\}.
    Return a JSON object with fields:
    \texttt{eq\_num},
    \texttt{eq\_transcriptions},
    \texttt{var\_definitions},
    \texttt{chain\_of\_thought},
    \texttt{math\_relevance}.\\

    \textbf{[INSERT PAGE]}\\[4pt]
    \rule{\linewidth}{0.4pt}\\[4pt]

    \textbf{System (Plot):}
    You are an expert researcher analyzing a simulation plot.
    Describe the plot in enough detail that a code generator can
    reproduce it exactly: subplot count, axis labels, units, curves,
    time range, expected qualitative behavior, and approximate
    numerical values.\\

    \textbf{User (Plot):}
    The image is a screenshot of a simulation plot.
    Problem: \{problem\_statement\}. Parameters: \{params\}.
    Describe the plot to guide downstream code generation.
    Do not overinterpret.
    Return a JSON object with fields:
    \texttt{subplot\_count},
    \texttt{subplot\_details},
    \texttt{time\_range},
    \texttt{plot\_behavior},
    \texttt{features},
    \texttt{parsed\_values}.\\

    \textbf{[INSERT SIMULATION PLOT FROM PAPER]}
  \end{tcolorbox}
  \caption{Prompt for Analyzer Agent.  The agent performs two tasks: transcribe red-boxed equations into readable format, analyze and describe behavior of the system using simulations from the paper. \textit{problem\_statement}, \textit{params} and \textit{init\_conditions} are defined by domain expert during annotation.}
  \label{fig:prompt-hint}
\end{figure}

\paragraph{Coder}

This agent uses the problem statement, parameter list, and paper context to write python code for the system using the proposed control method and initializing it with the provided parameters. The coder also writes a simulation function that matches the axes and input conditions of the target plot from the paper. The simulation function is executed using command line tools to save the generated plot, which is utilized for feedback. Figure~\ref{fig:prompt-coder} shows the instructions provided to the Coder agent.

\begin{figure}[!ht]
  \centering
  \small
  \begin{tcolorbox}[title=Agent 2: Coder --- Code Generation \& Repair]
    \textbf{System:}
    You are an expert control systems engineer and Python programmer.
    Output a single self-contained Python file.

    \textbf{User (Code Generation):}
    Problem: \{problem\_statement\}. Parameters: \{params\}.
    Paper analysis (equations + target plot description):
    \{paper\_context\}.\\
    Use the verification plot as a guide to understand the system ---
    do not hard-code values from it.
    Match the target plot layout (subplots, axes, grid, ticks).\\

    \rule{\linewidth}{0.4pt}\\[4pt]

    \textbf{User (Code Repair):}
    Problem: \{problem\_statement\}. Parameters: \{params\}.
    Paper analysis:
    \{paper\_context\}, Current candidate code: \{current\_code\}.\\
    Visual diagnosis from Feedback LLM: \{visual\_diagnosis\}.\\
    Fix the code so the generated plot matches the target.
    Cross-reference equation context to verify formulas.\\
  \end{tcolorbox}
  \caption{Prompt for Coder Agent.  The agent performs           both initial code generation and iterative code repair.  In repair
           mode, the current code and the feedback agent's visual
           diagnosis are appended to the user message.}
  \label{fig:prompt-coder}
\end{figure}

\paragraph{Verifier}

This agent compares the generated plot against the simulation plot from the paper. If the plots match visually, the loop terminates. If they deviate, the Verifier conducts a qualitative (simulation-level) and quantitative (code-level) analysis to generate corrective instructions. This feedback is routed back to the Coder to refine the implementation. We allow this generate-execute-verify loop to run for up to 8 iterations (set empirically). If the plots do not align after 8 attempts, the framework is terminated and the paper is marked as fail. The prompt is shown in Figure~\ref{fig:prompt-feedback}. 

\begin{figure}[!ht]
  \centering
  \small
  \begin{tcolorbox}[title=Agent 3: Verifier --- Visual Comparison]
    \textbf{System:}
    You are an expert control-systems engineer and data visualization
    specialist.  Compare simulation plots with meticulous attention to
    physical correctness.\\

    \textbf{User:}
    \textbf{Image 1} is the \textsc{generated} plot.
    \textbf{Image 2} is the \textsc{target} plot from the paper.\\

    Comparison criteria:
    (1)~Layout of sub-plots - same count?
    (2)~Axes and labels - same quantities?
    (3)~Qualitative curve shapes - same transient and steady-state?
    (4)~Approximate numerical ranges - peaks, settling values and time scales?

    Minor cosmetic differences are acceptable.  Focus on the underlying physics.\\

    If the plots match, end with:
    \texttt{MATCH\_CONFIRMED}.\\

    If they do not match, provide:
    (1)~Which sub-plot(s) differ and how;
    (2)~What curve behavior is wrong;
    (3)~Possible root-cause in the code;
    (4)~Specific mathematical fixes.\\

    \textbf{[INSERT GENERATED PLOT]}\\
    \textbf{[INSERT TARGET PLOT (FROM PAPER)]}
  \end{tcolorbox}
  \caption{Prompt for Verifier Agent:  The agent receives the generated and target plots side-by-side.  A confirmed match terminates the loop; otherwise a structured diagnosis feeds back into
  the Coder Agent for the next iteration.}
  \label{fig:prompt-feedback}
\end{figure}

\subsection{Evaluation Metrics}
\label{subsec:eval-metrics}

We define multi-axis metrics that assess both the simulation quality and code generation accuracy. Since ground truth code was absent in most of the papers studied, we define human and LLM-based metrics and study their correlation, providing insights in automated code analysis and fidelity.

\paragraph{Simulation-Level Metric} We use the \emph{Figure Reconstruction Score} (FRS), a score given by two independent domain experts and an LLM judge, to quantify the visual agreement between the generated and simulation plot. The score can be: (i) 1 (non-reconstructed): generated output fails to capture the behavior; (ii) 2 (partial reconstruction): some qualitative behavior is present, but key dynamics are missing/incorrect; (iii) 3 (near reconstruction): core behavior is recovered, with deviations due to initial conditions, scaling, or numerical differences; (iv) 4 (high-fidelity reconstruction): no visible differences from the reported figure.

\paragraph{Code-Level Metrics} We use two metrics to assess generated script. (i) Design Alignment Score (DA) measures how accurately (in \%) the  code  aligns with the  the paper and problem statement. (ii) Equation Coverage Score measures \% of equations implemented in the code. These scores are provided by the LLM after analyzing the  code, the paper context and the problem statement.


\begin{table}[h]
\centering
\caption{Qwen-3 paper filtering results across 500 candidate papers.}
\label{tab:qwen3_labels}
\begin{tabular}{lrc}
\toprule
\textbf{Label} & \textbf{Count} & \textbf{\% of Total} \\
\midrule
Recoverable       & 271 & 54.2\% \\
Not recoverable   & 120 & 24.0\% \\
Indecisive (no label)        & 109 &  21.8\% \\
\midrule
\textbf{Total}     & 500 & 100\%  \\
\bottomrule
\end{tabular}
\vspace{-10pt}
\end{table}

\begin{table}[b]
\centering
\setlength{\tabcolsep}{3.5pt}
\caption{One-pass vs.\ RESCORE across CDC 2021-2025 ($n = 194$).
         FRS Human is avg. of both raters.   Best values in a row are  \textbf{bold}.}
\label{tab:main_results}
\small
\begin{tabular}{l cc c cc c}
\toprule
& \multicolumn{3}{c}{\textbf{Single-Pass (FRS)}} & \multicolumn{3}{c}{\textbf{RESCORE (FRS)}} \\
\cmidrule(lr){2-4} \cmidrule(lr){5-7}
\textbf{Year ($n$)}
 & FRS-H & FRS-L & DA (\%)
 & FRS-H & FRS-L & DA (\%) \\
\midrule
2021 (38)  & 1.80 & 2.45 & 64.9 & \textbf{2.09} & \textbf{2.89} & \textbf{66.2} \\
2022 (41)  & 1.82 & 2.24 & 60.6 & \textbf{2.26} & \textbf{2.78} & \textbf{66.5} \\
2023 (46)  & 1.65 & 2.13 & 57.1 & \textbf{2.21} & \textbf{2.70} & \textbf{65.4} \\
2024 (40)  & 1.61 & 2.12 & 59.5 & \textbf{2.00} & \textbf{2.55} & \textbf{67.3} \\
2025 (29)  & 1.81 & 2.24 & 56.1 & \textbf{2.34} & \textbf{2.90} & \textbf{60.6} \\
\midrule
\textbf{All} & 1.73 & 2.23 & 59.7 & \textbf{2.17} & \textbf{2.75} & \textbf{65.5} \\
\bottomrule
\end{tabular}
\end{table}

\section{Results}
\label{sec:results}

We analyze the types of papers filtered by Qwen 3 for code recovarability. Then we quantitatively compare our method against a Single-Pass baseline where the LLM is provided with the expert annotations and simulation figures from the paper but is asked to generate the code without visual feedback. Lastly, we discuss  key findings. 


\subsection{LLM-Based Paper Filtering}\label{sec:filtering}

For each paper, Qwen-3 was tasked with producing a label
(\emph{recoverable} or \emph{not recoverable}) with justification based on the criteria defined in section~\ref{subsec:paper-selection-criteria}. Table~\ref{tab:qwen3_labels} summarizes
the outcomes.
Of the 500 papers, Qwen-3 assigned a definitive label to 391 and 69.3\% of them were labeled as
recoverable.

For the 120 papers classified as not recoverable, the primary rejection reasoning given by Qwen 3 was neural network training (25 papers, 20.3\%), no simulation figure (19 papers, 15.4\%) and dependence on external simulators (11 papers, 9.8\%). Some papers were rejected due to parameters cited from other references, lack of simulations (theoretical papers), or incomplete parameter specification. Filtering them allows us to test the hypothesis of automated code recovery based on the context within a paper. Most of the 109 unlabeled papers have ambiguous parameters or use domain-specific toolboxes. Qwen-3 is reasonably calibrated: rather than forcing a decision on ambiguous cases, it defers, which is preferable to systematic misclassification.  We conservatively exclude unlabeled papers from evaluation.

\subsection{Single-Pass vs. RESCORE Framework}\label{sec:main_results}

Table~\ref{tab:main_results} presents the simulation recoverability comparison across
all five CDCs. FRS Human (FRS-H) consists of the average scores annotated by two domain experts, and FRS LLM (FRS-L) shows the scores annotated by the LLM, comparing the generated simulation plot and the plot from the paper based on the criteria defined in section~\ref{subsec:eval-metrics}. Our framework consistently outperforms
single-pass generation on figure reconstruction fidelity and design alignment.
The mean FRS Human increases from
1.73 (single-pass) to 2.17 (RESCORE), a relative 
improvement of 25\%. We observed that the fraction of papers achieving near or high-fidelity reconstruction (FRS~$\geq3$) increases from 20.6\% to 32.0\%, indicating that the visual feedback is a crucial component to recovering better quality simulations from papers. 
The LLM grader, while  generous than human raters, reflects the same trend: mean FRS increases from 2.23$\rightarrow$2.75. Wilcoxon signed-rank test~\cite{wilcoxon} confirms significant FRS-H ($W=900$, $p < 0.001$, $r=0.64$) and FRS-L ($W=513$, $p < 0.001$, $r=0.73$) improvement across all 194 executed papers. Of the 100 papers whose human FRS changed with visual feedback, 74 improved under RESCORE while only 26 regressed, and the improvements were substantially larger in magnitude, collectively accounting for 82\% of the total ranked difference, indicating that visual feedback systematically rescues low-fidelity first-attempt reconstructions rather than producing sporadic gains.


\begin{table}[t]
\centering
\caption{Human annotation stats for Figure Reconstruction Score.}
\label{tab:human_eval}
\small
\begin{tabular}{lr}
\toprule
\textbf{Metric} & \textbf{Value} \\
\midrule
FRS Human 1 (mean)          & 2.05 \\
FRS Human 2 (mean)          & 2.29 \\
\midrule
\% papers FRS $\geq 3$ (H1 / H2) & 34.0\% / 41.2\% \\
\% papers FRS $= 4$ (H1 / H2)    & 19.6\% / 23.7\% \\
\midrule
Human inter-rater: exact     & 69.6\% \\
Human inter-rater: within-1  & 97.4\% \\
\bottomrule
\end{tabular}
\vspace{-10pt}
\end{table}

\begin{figure}[ht]
  \centering
  \includegraphics[width=\columnwidth]{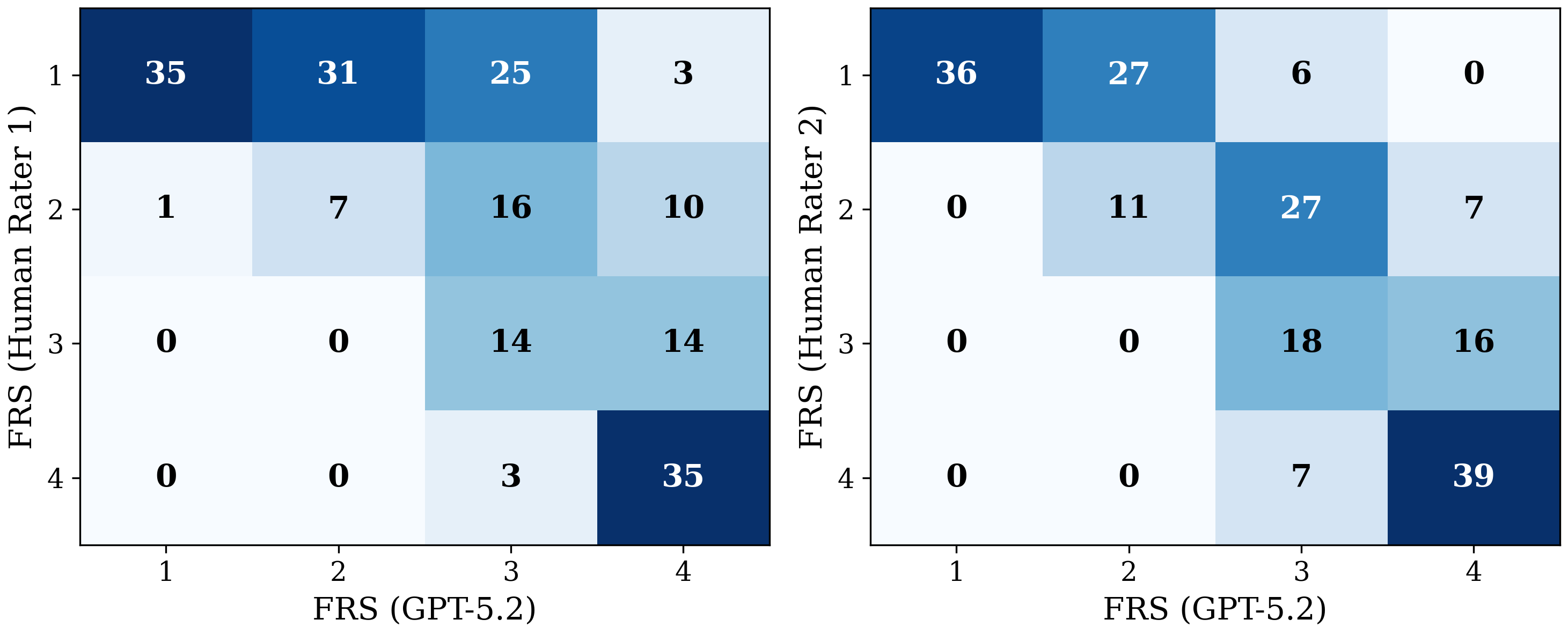}
  \caption{Confusion matrices: Human raters vs. LLM grader on RESCORE outputs. Off-diagonal mass above diagonal shows LLM optimism.}
  \label{fig:confusion}
  \vspace{-10pt}
\end{figure}

\begin{figure}[ht]
  \centering
  \includegraphics[width=\linewidth]{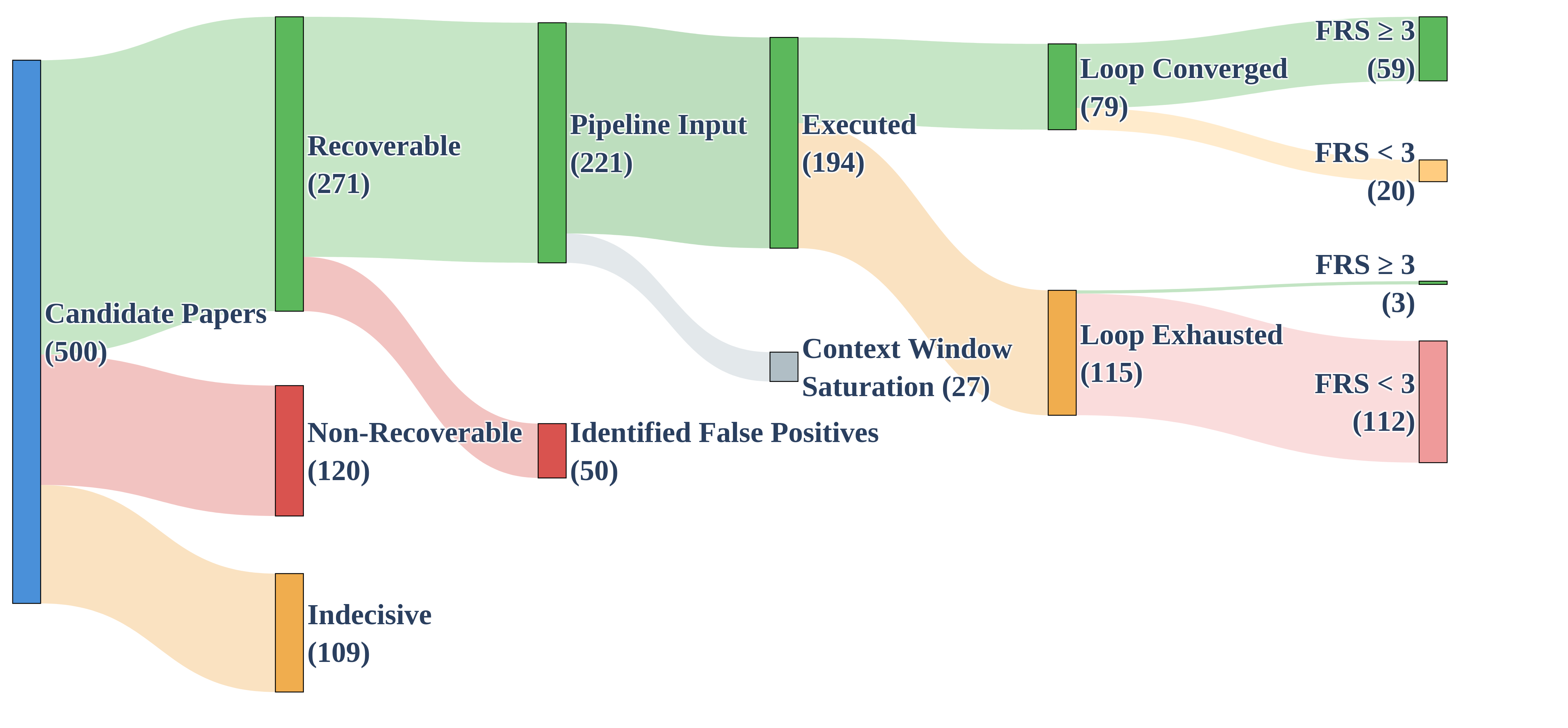}
  \caption{End-to-end attrition of the 500 candidate papers. Domain expert filtered some false positives. Of the 79 papers whose feedback loop converges, 59 (74.7\%) achieve FRS\,$\geq$\,3; among the 115 exhausted papers only 3 (2.6\%) do so; loop convergence  predicts reconstruction quality.}
  \label{fig:sankey}
\end{figure}

Table~\ref{tab:human_eval} summarizes the evaluation statistics of individual domain experts. The two raters agree exactly on 69.6\% of papers and are within one
score on 97.4\%, indicating strong inter-rater reliability on our proposed FRS ordinal scale. The automated LLM grader agrees exactly with Human~1 on 46.9\% of papers and with Human~2 on 53.6\%.  Within-one agreement is 80.4\% and 93.3\%, respectively.  Notably, the LLM assigns a higher score than the human in roughly half the cases, yielding a mean optimism bias of $+0.70$ over Human~1 and $+0.46$ over Human~2. This bias is concentrated at intermediate scores: when a human rates a paper FRS~$=1$, the LLM upgrades it to~2 or~3 in 59 of 94 cases (for Human~1), whereas at FRS~$=4$ the LLM agrees with the human 92\% of the time. See figure \ref{fig:confusion} for a correlation study between the LLM and human FRS graders. The LLM-based grader is a reliable filter for identifying successes and failures, but tends to over-credit partial reconstructions where a domain expert may recognize qualitative mismatches in the dynamics. Figure~\ref{fig:sankey} shows the path from our initial 500 paper corpus to the final analyzed papers. Of the 221 papers that passed human review, 27 were excluded
because their input prompts exceeded the model's context window.
While GPT-5.2 provides a 128K-token context, tokenized text input in RESCORE comprises the system prompt, the synthesised paper context (parsed equations and plot
descriptions), and high-detail screenshot of the target figure consuming
roughly 1,500 tokens.  Papers with dense mathematical content
(e.g.\ long derivations spanning five or more pages) can saturate the input budget before code generation begins.
We chose a fixed context budget for all papers to ensure a fair
and cost-controlled evaluation rather than selectively expanding
the window for individual papers. RESCORE requires approximately \$2.5 for each paper, a cost that will reduce with the rapid advancement of LLMs.

Beyond simulation plot-based metrics, we notice improvement on the design alignment score, which measures the quality of code generated and if it faithfully answers the expert-annotated problem statement. Due to the scale and diversity of papers included in our study, we cannot manually verify the quality of code, which often requires domain expertise. To mitigate bias in our scoring, we use two LLMs (GPT-5.2 and Gemini 3.1-flash-lite \cite{gemini31flashlite_modelcard2026}) to evaluate the generated code and report an average score. Our framework improves design alignment scores across all five years, reflecting a clear message: simulation plots are strong steering signals for agentic systems and can align the code-generation capabilities of modern LLMs for complex control system design. While we also ask the LLM to provide an equation coverage metric, this value is similar across both single-pass (72.8\%) and our framework (74.3\%), indicating that feedback from simulation plots is difficult to correlate with missing mathematical equations in the generated code. 




\subsection{The Role of Visual Feedback}\label{sec:feedback}

\begin{figure}[ht]
  \centering
  \includegraphics[width=\columnwidth]{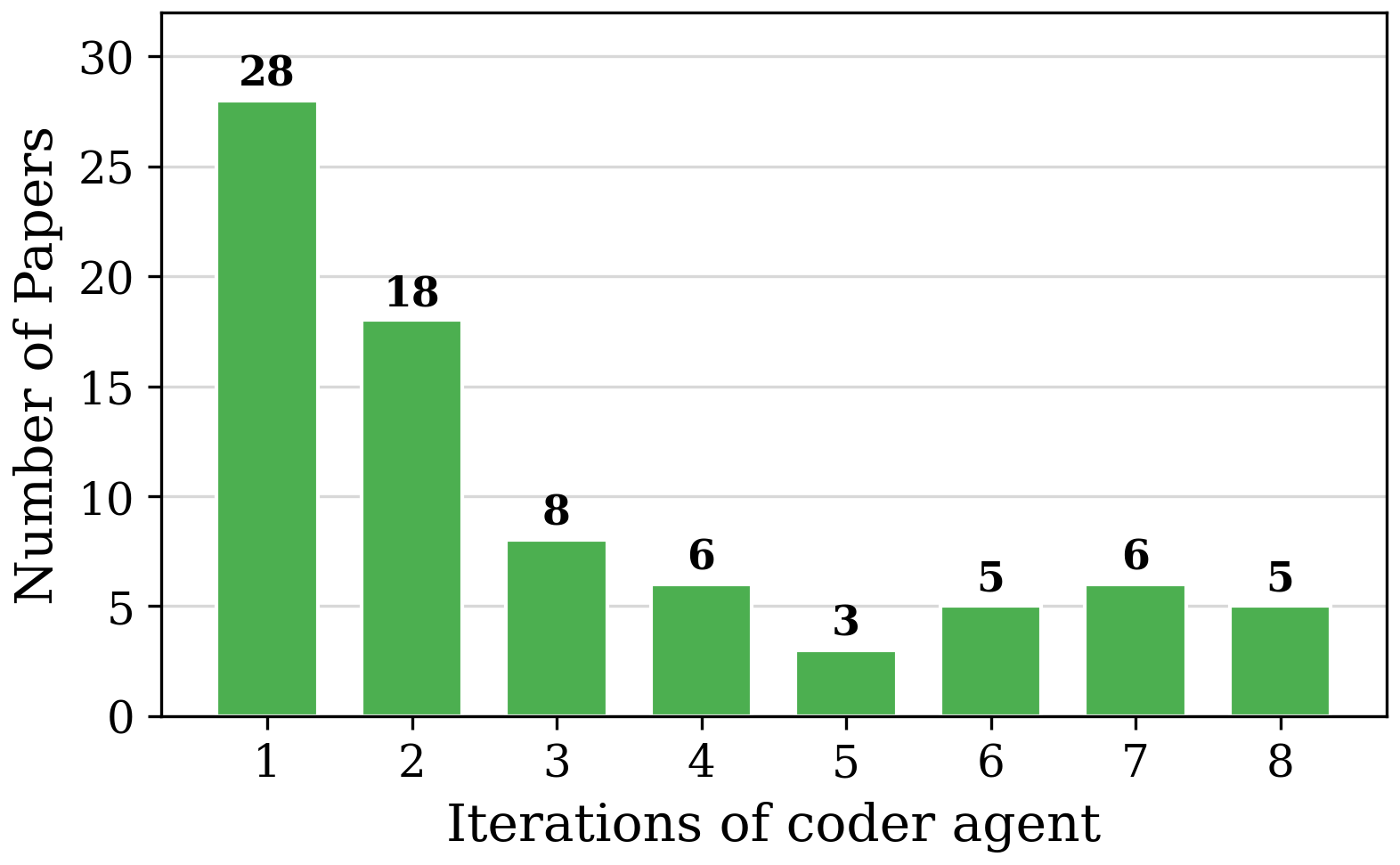}
  \vspace*{-0.25in}
  \caption{Distribution of papers for which the RESCORE loop converged and the iterations of coder agent needed to recover the simulation code.}
  \label{fig:iterations}
  
\end{figure}

Figure~\ref{fig:iterations} shows the distribution of feedback
iterations of the 79 papers that converged after the RESCORE loop. Most converge within 1--3 iterations, suggesting  most errors are local: sign flips in dynamics, wrong initial conditions, or missing terms in a control law can be diagnosed by a single visual comparison. 
A long tail of 7–8 iterations is common in multi-subsystem simulations (e.g., observer-controller-plant cascades), where each round fixes one deficiency but exposes another as modules must be debugged  sequentially. These 79 papers represent the 40.7\% of 194 papers for which our framework recovered task-coherent simulations.

\vspace{-10pt}
\subsection{Year-over-Year Trends}\label{sec:year}

In Table~\ref{tab:main_results},  mean FRS Human ranges from 1.61 (CDC~2024, single-pass) to 2.34
(CDC~2025, RESCORE) with no monotonic trend across years.
RESCORE improves every year, ranging from
$+0.29$ (2021) to $+0.56$ (2023).
The absence of a strong year effect suggests that the difficulty of simulation recovery is determined by the mathematical structure of a paper, type of dynamics, complexity of controller synthesis, and number of interacting subsystems. 


\subsection{Qualitative Analysis of RESCORE}%
\label{sec:qualitative}
\paragraph{Common failure modes of RESCORE}
The most frequent cause of failure is
\emph{optimization-problem substitution}: the LLM replaces the paper's formulation (sum-of-squares programs, linear matrix inequalities (LMIs), mixed-integer linear programs (MILPs), or semi-definite programs (SDPs)) with a simpler surrogate such as a QP or heuristic.
This pattern is prevalent in data-driven SDP-based control and signal temporal logic (STL) specification papers, where the core contribution is the optimization formulation. Although visual feedback partially mitigates this for some cases, the substitution remains the single largest source of initial failure, with these papers recording the lowest single-pass FRS (1.50) Another recurring failure is \emph{incomplete PDE discretization}: papers involving spatially-distributed systems (reaction--diffusion equations, hyperbolic PDEs, backstepping kernels) receive only truncated modal approximations that do not capture the full spatial dynamics. These papers show
\emph{no improvement} under feedback ($\Delta = -0.07$); none of the
four PDE papers starting at FRS\,$=$\,1 are rescued, because truncated spatial discretizations produce outputs that appear
qualitatively plausible yet are quantitatively wrong. Third, for multi-agent and networked systems, the model simplifies the communication topology or replaces distributed protocols with centralized ones. These papers exhibit a similarly
muted response ($\Delta = +0.09$, rescue rate 5.9\%), as topology simplification or protocol centralization generates plots that are visually similar but dynamically incorrect.


\begin{table*}[ht]
  \centering
  \caption{Examples of visual-feedback improvement.  Each row shows ground-truth 
    (\emph{Target}), LLM's first-attempt output before any
    feedback (\emph{Single-pass}), and output after the
    RESCORE feedback loop (\emph{RESCORE}).
    Human FRS scores are beneath each plot.}
  \label{tab:qualitative}
  \setlength{\tabcolsep}{4pt}
  \renewcommand{\arraystretch}{1.2}
  \begin{tabular}{>{\centering\arraybackslash}m{0.31\textwidth}
                  >{\centering\arraybackslash}m{0.31\textwidth}
                  >{\centering\arraybackslash}m{0.31\textwidth}}
    \toprule
    \textbf{Target} & \textbf{Single-pass} & \textbf{RESCORE} \\
    \midrule
    \multicolumn{3}{c}{\small\textit{Data-Driven Reachability with Christoffel Functions~\cite{devonport_cdc2021}}} \\[2pt]
    \includegraphics[width=\linewidth]{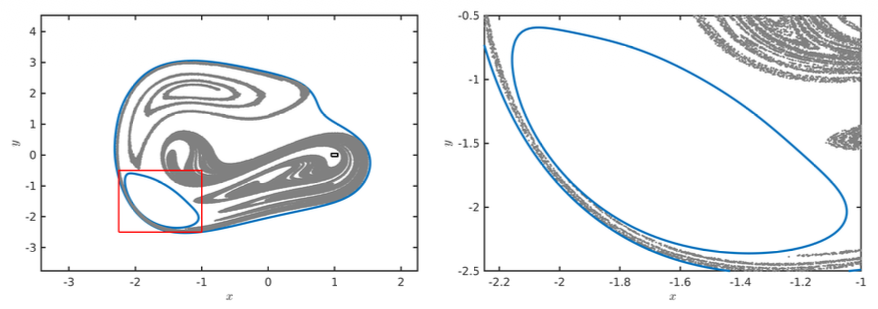}
    \newline {\scriptsize Plot from paper}
    &
    \includegraphics[width=\linewidth]{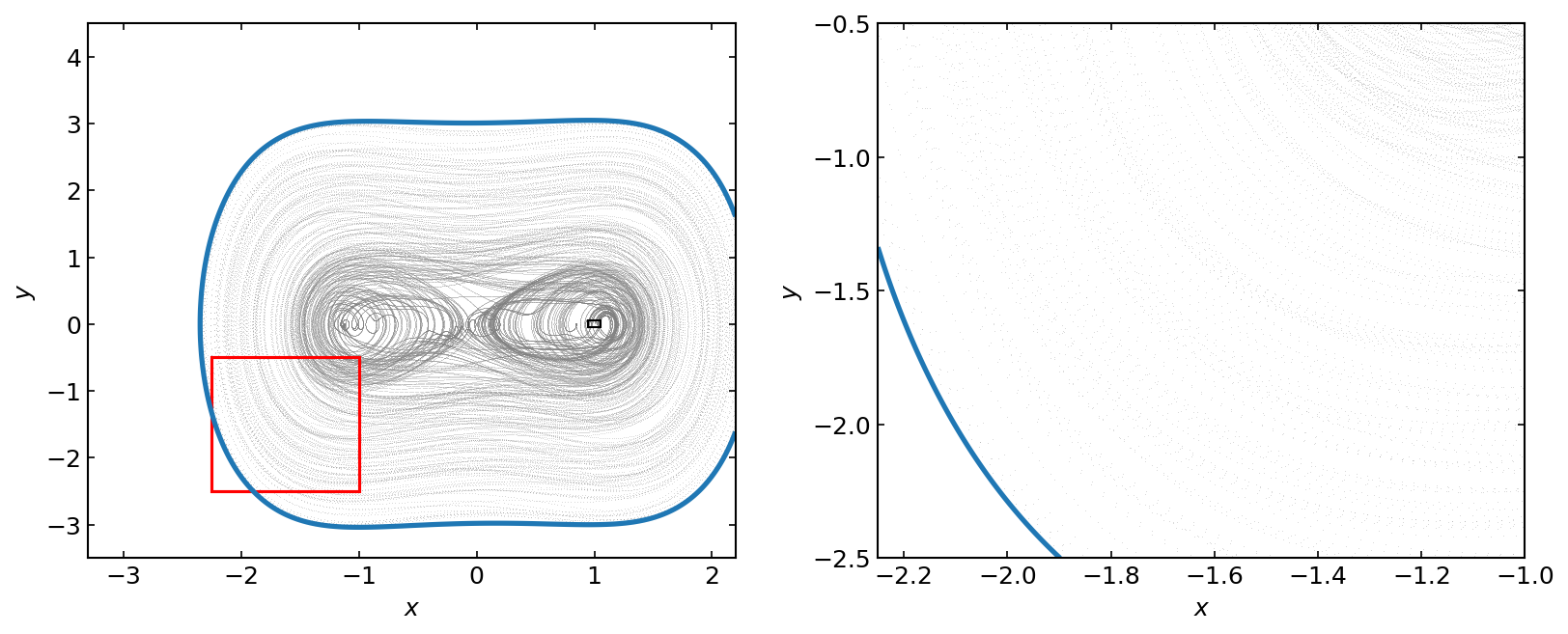}
    \newline {\scriptsize FRS\,=\,1.0}
    &
    \includegraphics[width=\linewidth]{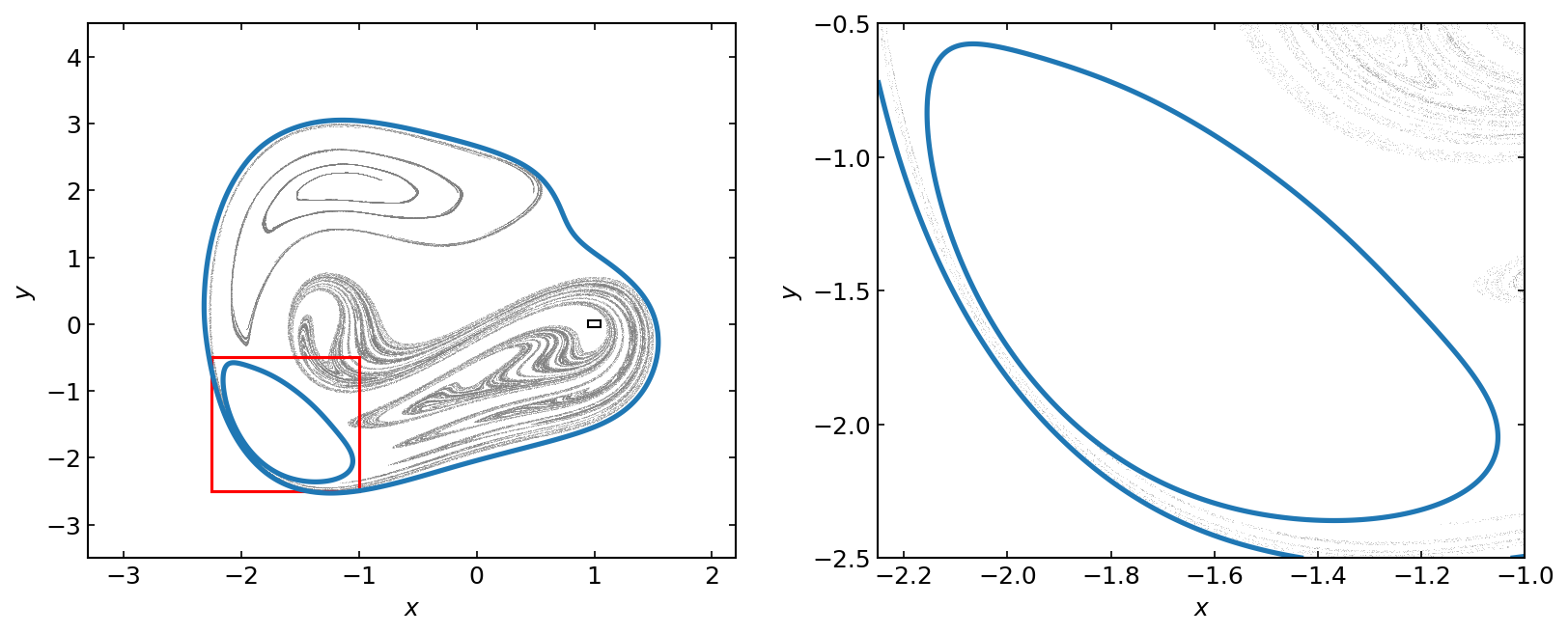}
    \newline {\scriptsize FRS\,=\,4.0}
    \\[6pt]
    \midrule
    \multicolumn{3}{c}{\small\textit{Identification of Piecewise Affine Systems with Online Deterministic Annealing~\cite{mavridis_cdc2023}}} \\[2pt]
    \includegraphics[width=\linewidth]{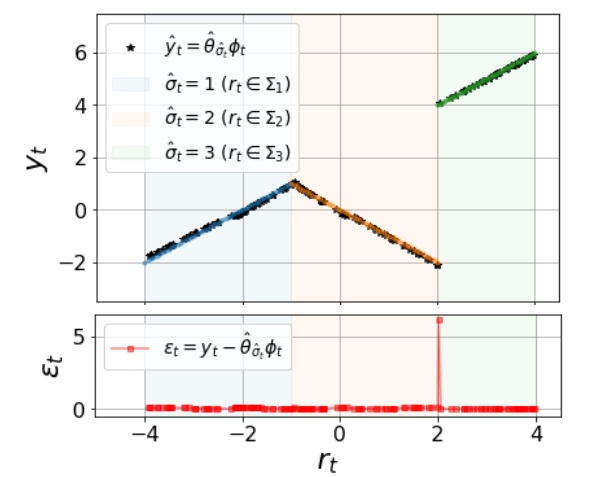}
    \newline {\scriptsize Plot from paper}
    &
    \includegraphics[width=\linewidth]{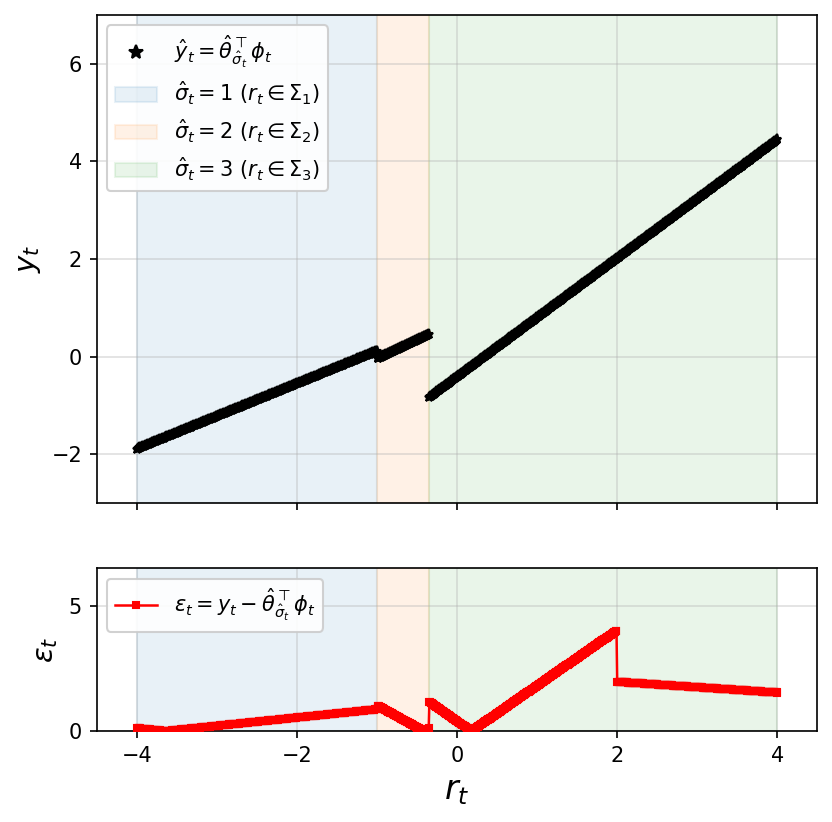}
    \newline {\scriptsize FRS\,=\,2.0}
    &
    \includegraphics[width=\linewidth]{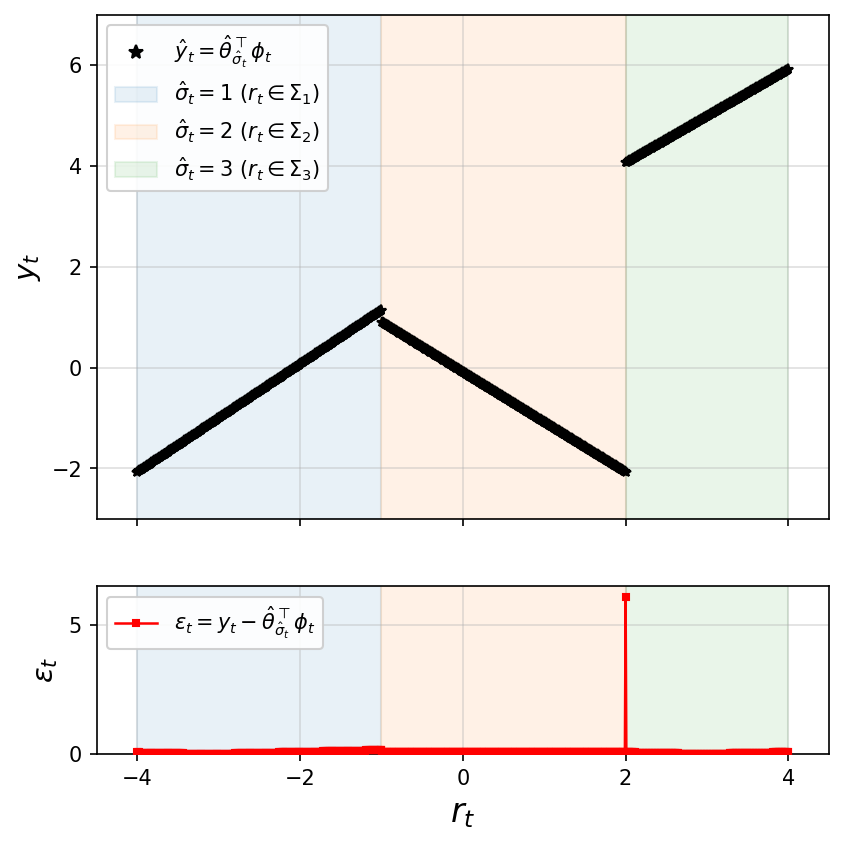}
    \newline {\scriptsize FRS\,=\,4.0}
    \\[6pt]
    \bottomrule
  \end{tabular}
  \vspace{-10pt}
\end{table*}

\paragraph{What does RESCORE feedback loop correct?} Analysis of papers that improve from FRS~$=1$ (Single-Pass) to FRS~$\geq3$ (RESCORE) reveals that the corrections are mostly local: adjusting a sign in a feedback gain, fixing an initial condition, correcting a discretization time-step, or adding a missing term in the plant dynamics. Visual comparison is effective at diagnosing such errors. For example, if the generated response is overdamped but the target shows oscillatory behavior, the Verifier can implicate a missing imaginary-axis eigenvalue in the closed-loop system. This suggests that for control system simulations, where qualitative shape of the response is tied to the mathematical structure, visual feedback offers high information density diagnostic. Table~\ref{tab:qualitative} shows examples where visual feedback improves  simulation.

\begin{figure}[ht]
  \centering
  \includegraphics[width=\columnwidth]{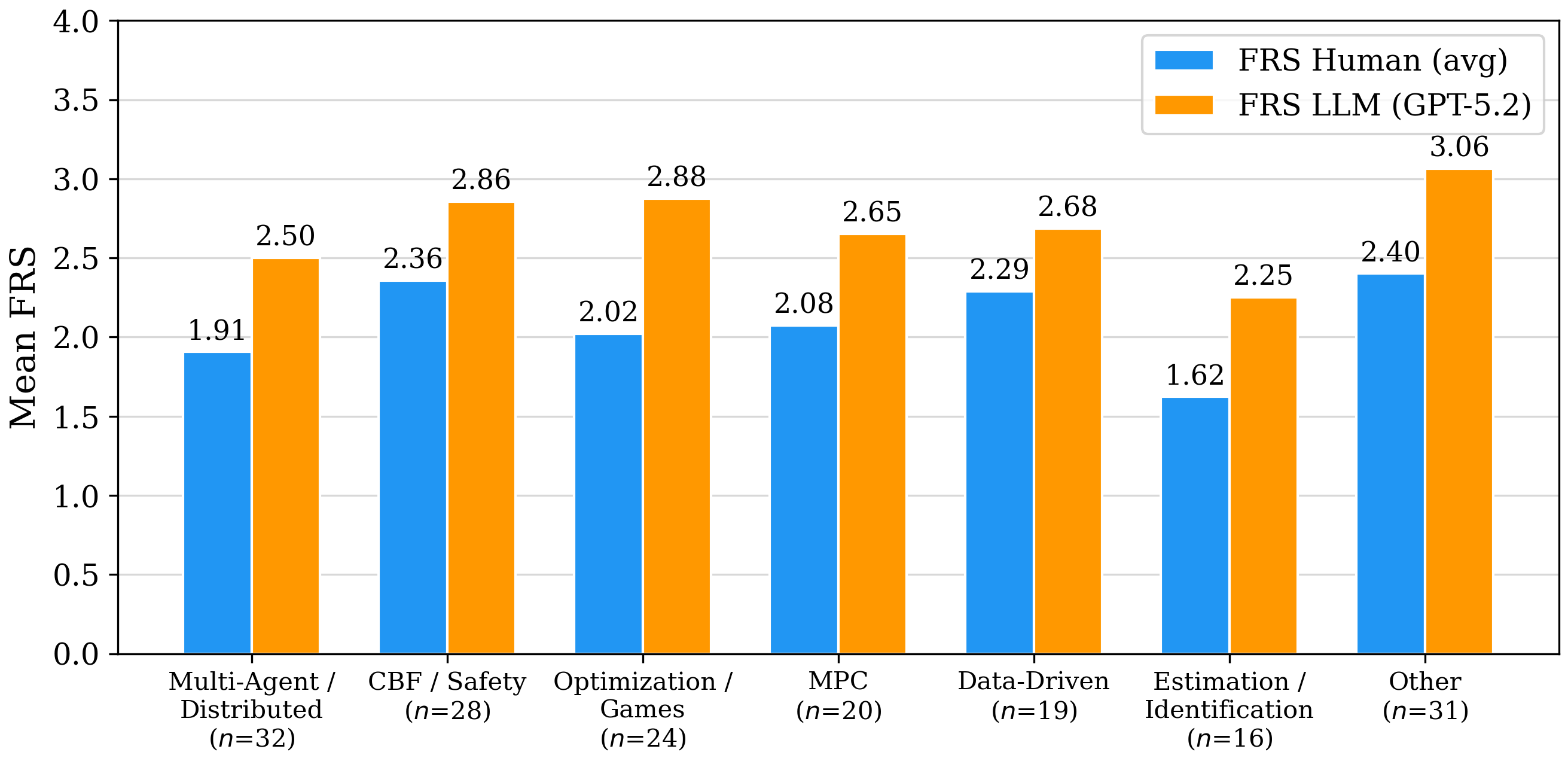}
  \caption{Mean FRS by category (RESCORE, $n=194$).
           Categories assigned by keyword match of paper
           titles. Human FRS avg. of both raters.}
  \label{fig:category_frs}
  \vspace{-15pt}
\end{figure}

\paragraph{Performance by  Category}\label{sec:categories}
Figure~\ref{fig:category_frs} breaks down FRS by paper category,
where each paper is assigned a topic via keyword matching on its title. 
CBF/Safety papers have the highest human FRS (2.36), 
because control barrier function formulations produce well-structured
simulation loops. 
Estimation/Identification papers score lowest (1.62), reflecting the
difficulty of reproducing observer dynamics and filter tuning without
explicit numerical specifications.

\subsection{Speed-Up Compared with Human Replication}
For a graduate-level researcher without domain expertise, deciphering the mathematics, resolving implicit numerical choices, and iteratively debugging code to match a published figure can demand on the order of 10--20 hours of effort~\cite{Gligoric2024In}. In contrast, RESCORE framework completes end-to-end extraction, coding, and visual verification loop in about 60--90 minutes. When successful (FRS $\geq 3$), it yields $\sim$10$\times$ speed-up over manual replication. Even for partial reconstructions, the codebase is a structurally sound foundation and can reduce the effort to complete the simulation.

\begin{figure*}[t]
  \centering
  \includegraphics[width=\textwidth]{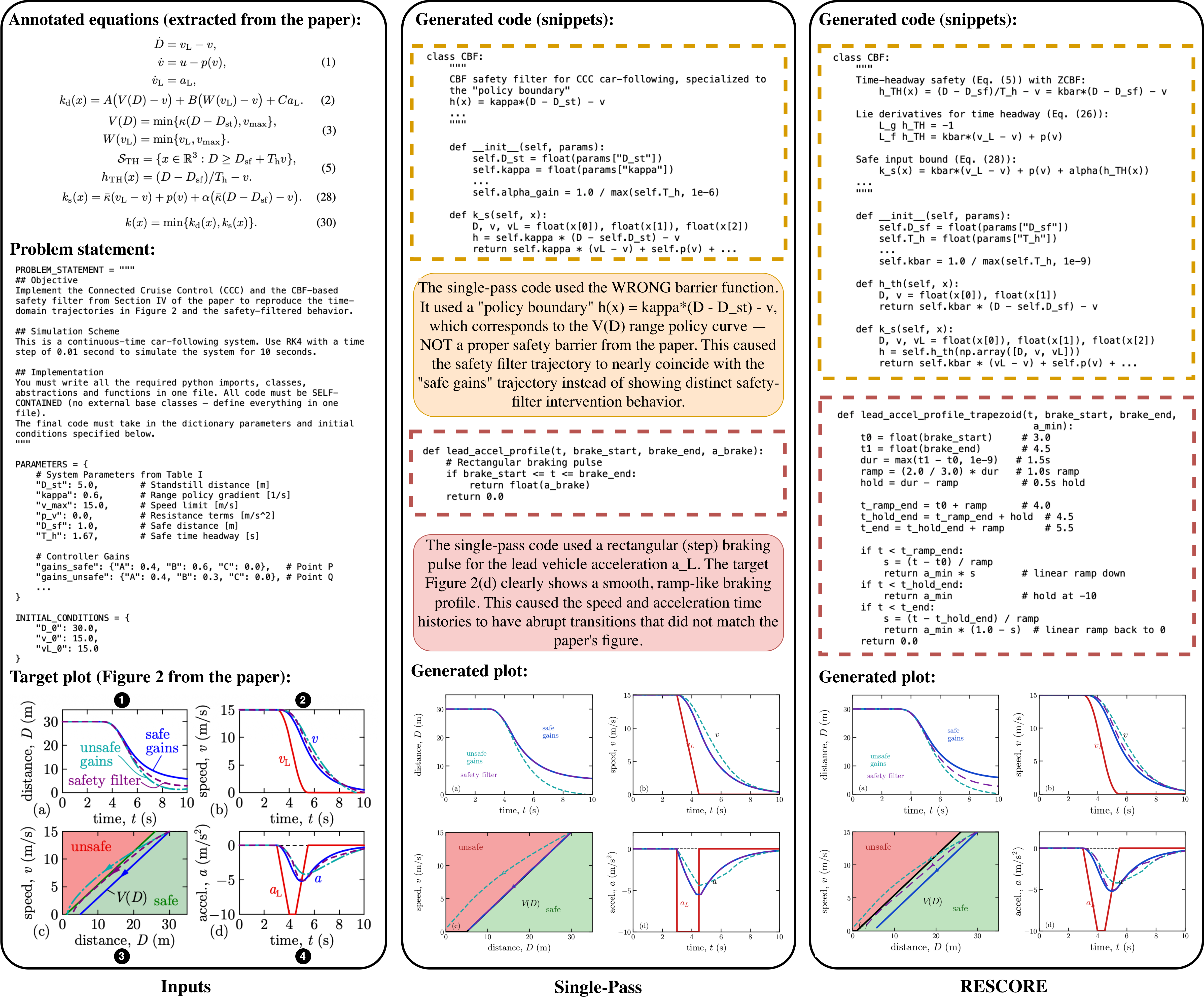}
  \caption{Case study for Connected Cruise Control system from \cite{molnar2023safety}. \textbf{Inputs (Left):} Agentic pipeline is provided with expert-annotated equations, problem statement, and target simulation Figure 2 from \cite{molnar2023safety}. \textbf{Single-Pass (Center):} The simulation code compiles without syntax errors but has semantic flaws, such as implementing an incorrect control barrier function and a rigid step-braking profile, resulting in visual discrepancies. \textbf{(Right):} Guided by visual feedback, the {\bf RESCORE} closed-loop diagnoses mismatches, corrects underlying mathematical structure, reconstructing the simulation.}
  \label{fig:case_study_1}
  \vspace{-10pt}
\end{figure*}

\subsection{Case Study}
As a case study, consider the paper \cite{molnar2023safety} from CDC 2023 for an example (see Fig.~\ref{fig:case_study_1}). This work defines a Connected Cruise Control (CCC) system for a car-following scenario, proposing a CBF based safety filter to minimally modify nominal inputs to guarantee safety. The automated pipeline was tasked with reproducing Figure 2 from the paper, which is a 2x2 subplot detailing distance, speed, phase portraits, and acceleration under safe gains, unsafe gains, and safety-filtered control. As shown in the ``Inputs" column of the accompanying figure, the LLM agent was provided with the problem statement, initial conditions, and screenshots of the governing equations. The pipeline completed the generation in two iterations, demonstrating the use of visual feedback.

Single-Pass generated code compiled and executed without syntax errors, successfully implementing the plant dynamics and nominal CCC control law. However, visual comparison against the target plot revealed severe discrepancies caused by three distinct errors. (i) \textbf{Incorrect barrier function:} the model conflated two structurally similar equations. It implemented the ``policy boundary'' $h(x) = \kappa(D - D_{\text{st}}) - v$ instead of the correct time-headway safety barrier $h_{\text{TH}}(x) = (D - D_{\text{sf}})/T_h - v$ (see target plot \circlethree{} in figure~\ref{fig:case_study_1}). Since the numerical slopes were similar ($\kappa = 0.6$ vs $\bar{\kappa} = 1/T_h \approx 0.599$), the mistake was subtle, but the difference in intercepts ($D_{\text{st}} = 5.0$ vs $D_{\text{sf}} = 1.0$) caused the filter to act too aggressively. As a result, the generated safety filter trajectory improperly coincided with the ``safe gains'' trajectory, failing to show the distinct intervention behavior visible in the paper. (ii) \textbf{Inaccurate braking profile:} Target plot \circlefour{} showed a smooth, trapezoidal ramp-hold-ramp braking profile for the lead vehicle. The initial code modeled this as a sudden, rectangular step pulse. This caused the lead vehicle's speed to drop to zero prematurely, resulting in abrupt, incorrect transitions in the downstream acceleration and speed plots. (iii) \textbf{Wrong phase portrait boundary:} For the phase plane plot \circlethree{},  the code incorrectly plotted the range policy $V(D)$ as the boundary separating the ``safe'' and ``unsafe'' regions, rather than the time-headway boundary from the CBF formulation. This misrepresented the region where the safety filter prevents constraint violations.

RESCORE diagnosed and fixed these issues. The verifier agent compared the single pass generated plot with the target, recognizing that the safety filter behavior was missing and that the lead vehicle's acceleration profile was modeled as an abrupt step function rather than a trapezoidal ramp. During the second iteration, these insights guided the coder agent. The code was updated to use the correct $D_{\text{sf}}$ and $\bar{\kappa} = 1/T_h$ for the time-headway barrier, a linear ramp-down function was introduced for the braking profile, and the phase portrait boundary was mapped correctly. As seen in the ``RESCORE" column, revised code produced a high-fidelity match with target plot. Visual feedback catches physically meaningful errors such as conceptually different but algebraically similar boundaries that standard syntactic tests would miss.

\section{Conclusion}

RESCORE reconstructed simulations for 40.7\% of evaluated papers, achieving a 25\% relative improvement over single-pass generation and an estimated 10$\times$ speed-up over manual replication. However, only 44.2\% of 500 recent CDC papers had sufficient parameter specifications for recovery, underscoring the need for stricter reporting standards. Remaining limitations include advanced optimization formulations (SDPs, LMIs) and PDE discretizations, where visually plausible plots can mask quantitative errors. We will open-source the benchmark, prompts, and evaluation pipeline to support reproducible control systems research.

\bibliographystyle{IEEEtran}
\bibliography{Master_Paper2Simulation}

@inproceedings{badertdinov2025swe,
	address = {(San Diego, CA)},
	author = {Badertdinov, Ibragim and Golubev, Alexander and Nekrashevich, Maksim and Shevtsov, Anton and Karasik, Simon and Andriushchenko, Andrei and Trofimova, Maria and Litvintseva, Daria and Yangel, Boris},
	booktitle = {Proc. Conf. on Neural Info. Processing Systems},
	month = {Dec},
	title = {SWE-rebench: An Automated Pipeline for Task Collection and Decontaminated Evaluation of Software Engineering Agents},
	year = {2025}}

@misc{gemini31flashlite_modelcard2026,
	author = {{Google DeepMind}},
	month = {March},
	title = {Gemini 3.1 Flash-Lite - Model Card},
	url = {https://deepmind.google/models/model-cards/gemini-3-1-flash-lite/},
	year = {2026}}

@inproceedings{molnar2023safety,
	address = {(Marina Bay Sands, Singapore)},
	author = {Molnar, Tamas G and Orosz, G{\'a}bor and Ames, Aaron D},
	booktitle = {Proc. IEEE Conf. on Decision and Control},
	month = {Dec},
	title = {On the safety of connected cruise control: analysis and synthesis with control barrier functions},
	year = {2023}}

@misc{openai2025gpt52,
	author = {{OpenAI}},
	title = {Introducing {GPT-5.2}: Announcement and System Card},
	url = {https://openai.com/index/introducing-gpt-5-2/},
	year = {2025}}

@article{yang2025qwen3,
	author = {Yang, An and Li, Anfeng and Yang, Baosong and Zhang, Beichen and Hui, Binyuan and Zheng, Bo and Yu, Bowen and Gao, Chang and Huang, Chengen and Lv, Chenxu and others},
	journal = {arXiv preprint arXiv:2505.09388},
	title = {Qwen3 technical report},
	year = {2025}}

@inproceedings{gouc2024ritic,
	address = {(Vienna, Austria)},
	author = {Gou, Zhibin and Shao, Zhihong and Gong, Yeyun and Yang, Yujiu and Duan, Nan and Chen, Weizhu and others},
	booktitle = {Proc. International Conf. on Learning Representations},
	month = {May},
	title = {CRITIC: Large Language Models Can Self-Correct with Tool-Interactive Critiquing},
	year = {2024}}

@inproceedings{yao2023react,
	address = {(Kigali, Rwanda)},
	author = {Yao, Shunyu and Zhao, Jeffrey and Yu, Dian and Du, Nan and Shafran, Izhak and Narasimhan, Karthik R and Cao, Yuan},
	booktitle = {Proc. International Conf. on Learning Representations},
	month = {May},
	title = {React: Synergizing reasoning and acting in language models},
	year = {2023}}

@inproceedings{gehring2025rlef,
	address = {(Vancouver, Canada)},
	author = {Gehring, Jonas and Zheng, Kunhao and Copet, Jade and Mella, Vegard and Cohen, Taco and Synnaeve, Gabriel},
	booktitle = {Proc. International Conf. on Machine Learning},
	month = {July},
	title = {RLEF: Grounding Code LLMs in Execution Feedback with Reinforcement Learning},
	year = {2025}}

@inproceedings{chen2024teaching,
	address = {(Vienna, Austria)},
	author = {Chen, Xinyun and Lin, Maxwell and Sch{\"a}rli, Nathanael and Zhou, Denny},
	booktitle = {Proc. International Conf. on Learning Representations},
	month = {May},
	title = {Teaching Large Language Models to Self-Debug},
	year = {2024}}

@inproceedings{du2024evaluating,
	address = {(Lisbon, Portugal)},
	author = {Du, Xueying and Liu, Mingwei and Wang, Kaixin and Wang, Hanlin and Liu, Junwei and Chen, Yixuan and Feng, Jiayi and Sha, Chaofeng and Peng, Xin and Lou, Yiling},
	booktitle = {Proc. IEEE/ACM International Conf. on Software Engineering},
	month = {April},
	title = {Evaluating large language models in class-level code generation},
	year = {2024}}

@inproceedings{liu2024repobench,
	address = {(Vienna, Austria)},
	author = {Liu, Tianyang and Xu, Canwen and McAuley, Julian},
	booktitle = {Proc. International Conf. on Learning Representations},
	month = {May},
	title = {RepoBench: Benchmarking Repository-Level Code Auto-Completion Systems},
	year = {2024}}

@inproceedings{yang2024swe,
	address = {(Vancouver, Canada)},
	author = {Yang, John and Jimenez, Carlos E and Wettig, Alexander and Lieret, Kilian and Yao, Shunyu and Narasimhan, Karthik and Press, Ofir},
	booktitle = {Proc. Conf. on Neural Information Processing Systems},
	month = {Dec},
	title = {Swe-agent: Agent-computer interfaces enable automated software engineering},
	year = {2024}}

@inproceedings{jimenez2024swe,
	address = {(Vienna, Austria)},
	author = {Jimenez, Carlos E and Yang, John and Wettig, Alexander and Yao, Shunyu and Pei, Kexin and Press, Ofir and Narasimhan, Karthik R},
	booktitle = {Proc. International Conf. on Learning Representations},
	month = {May},
	title = {SWE-bench: Can Language Models Resolve Real-world Github Issues?},
	year = {2024}}

@inproceedings{yu2025utboost,
	address = {(Vienna, Austria)},
	author = {Yu, Boxi and Zhu, Yuxuan and He, Pinjia and Kang, Daniel},
	booktitle = {Proc. Annual Meeting of the Assoc. for Computational Linguistics},
	month = {July},
	title = {Utboost: Rigorous evaluation of coding agents on swe-bench},
	year = {2025}}

@article{bhat_brainbodyllm,
author = {Bhat, Vineet and Kaypak, Ali Umut and Krishnamurthy, Prashanth and Karri, Ramesh and Khorrami, Farshad},
title = {Grounding Large Language Models for Robot Task Planning Using Closed-Loop State Feedback},
journal = {Advanced Robotics Research},
volume = {},
number = {},
pages = {},
doi = {},
url = {},
eprint = {},
year = {2025}
}

@article{li2022competition,
	author = {Li, Yujia and Choi, David and Chung, Junyoung and Kushman, Nate and Schrittwieser, Julian and Leblond, R{\'e}mi and Eccles, Tom and Keeling, James and Gimeno, Felix and Dal Lago, Agustin and others},
	journal = {Science},
	number = {6624},
	pages = {1092--1097},
	publisher = {American Association for the Advancement of Science},
	title = {Competition-level code generation with alphacode},
	volume = {378},
	year = {2022}}

@inproceedings{hendrycks2021measuring,
	address = {(Virtual)},
	author = {Hendrycks, Dan and Basart, Steven and Kadavath, Saurav and Mazeika, Mantas and Arora, Akul and Guo, Ethan and Burns, Collin and Puranik, Samir and He, Horace and Song, Dawn and others},
	booktitle = {Proc. Conf. on Neural Inf. Processing Systems},
	month = {Dec.},
	title = {Measuring coding challenge competence with apps},
	year = {2021}}

@article{chen2021evaluating,
	author = {Chen, Mark and Tworek, Jerry and Jun, Heewoo and Yuan, Qiming and Pinto, Henrique Ponde De Oliveira and Kaplan, Jared and Edwards, Harri and Burda, Yuri and Joseph, Nicholas and Brockman, Greg and others},
	journal = {arXiv preprint arXiv:2107.03374},
	title = {Evaluating large language models trained on code},
	year = {2021}}

@book{national2019reproducibility,
	author = {National Academies of Sciences and Medicine and Policy and Global Affairs and Board on Research Data and Information and Division on Engineering and Physical Sciences and Committee on Applied and Theoretical Statistics and others},
	publisher = {National Academies Press},
	title = {Reproducibility and replicability in science},
	year = {2019}}

@article{austin2021program,
	author = {Austin, Jacob and Odena, Augustus and Nye, Maxwell and Bosma, Maarten and Michalewski, Henryk and Dohan, David and Jiang, Ellen and Cai, Carrie and Terry, Michael and Le, Quoc and others},
	journal = {arXiv preprint arXiv:2108.07732},
	title = {Program synthesis with large language models},
	year = {2021}}

@article{blecher2023nougat,
	author = {Blecher, Lukas and Cucurull, Guillem and Scialom, Thomas and Stojnic, Robert},
	journal = {arXiv preprint arXiv:2308.13418},
	title = {Nougat: Neural optical understanding for academic documents},
	year = {2023}}

@inproceedings{deng2017image,
	address = {(Sydney, Australia)},
	author = {Deng, Yuntian and Kanervisto, Anssi and Ling, Jeffrey and Rush, Alexander M},
	booktitle = {Proc. International Conf. on Machine Learning},
	month = {August},
	title = {Image-to-markup generation with coarse-to-fine attention},
	year = {2017}}

@inproceedings{lopez2010humb,
	address = {(Uppsala, Sweden)},
	author = {Lopez, Patrice and Romary, Laurent},
	booktitle = {Proc. International Workshop on Semantic Evaluation},
	month = {July},
	title = {HUMB: Automatic key term extraction from scientific articles in GROBID},
	year = {2010}}

@article{shen2022vila,
	author = {Shen, Zejiang and Lo, Kyle and Wang, Lucy Lu and Kuehl, Bailey and Weld, Daniel S and Downey, Doug},
	journal = {Transactions of the Association for Computational Linguistics},
	pages = {376--392},
	publisher = {MIT Press One Broadway, 12th Floor, Cambridge, Massachusetts 02142, USA~{\ldots}},
	title = {VILA: Improving structured content extraction from scientific PDFs using visual layout groups},
	volume = {10},
	year = {2022}}

@inproceedings{clark2016pdffigures,
	address = {(Newark, NJ)},
	author = {Clark, Christopher and Divvala, Santosh},
	booktitle = {Proc. ACM/IEEE-CS Joint Conf. on Digital Libraries},
	month = {July},
	title = {Pdffigures 2.0: Mining figures from research papers},
	year = {2016}}

@inproceedings{seo2025paper2code,
	address = {(Rio de Janeiro, Brazil)},
	author = {Seo, Minju and Baek, Jinheon and Lee, Seongyun and Hwang, Sung Ju},
	booktitle = {Proc. International Conf. on Learning Representations},
	month = {April},
	title = {Paper2code: Automating code generation from scientific papers in machine learning},
	year = {2026}}

@article{starace2025paperbench,
	author = {Starace, Giulio and Jaffe, Oliver and Sherburn, Dane and Aung, James and Chan, Jun Shern and Maksin, Leon and Dias, Rachel and Mays, Evan and Kinsella, Benjamin and Thompson, Wyatt and others},
	journal = {arXiv preprint arXiv:2504.01848},
	title = {PaperBench: Evaluating AI's Ability to Replicate AI Research},
	year = {2025}}

@article{how2018control,
	author = {How, Jonathan P},
	journal = {IEEE Control Systems Magazine},
	number = {4},
	pages = {3--4},
	publisher = {IEEE},
	title = {Control systems reproducibility challenge [from the editor]},
	volume = {38},
	year = {2018}}

@inproceedings{devonport_cdc2021,
	address = {(Austin, TX)},
	author = {Devonport, Alex and Yang, Forest and El Ghaoui, Laurent and Arcak, Murat},
	booktitle = {Proc. IEEE Conf. on Decision and Control},
	month = {Dec},
	title = {Data-Driven Reachability Analysis with Christoffel Functions},
	year = {2021}}

@inproceedings{mavridis_cdc2023,
	address = {(Marina Bay Sands, Singapore)},
	author = {Mavridis, Christos N. and Baras, John S.},
	booktitle = {Proc. IEEE Conf. on Decision and Control},
	month = {Dec},
	title = {Identification of Piecewise Affine Systems with Online Deterministic Annealing},
	year = {2023}}

@inproceedings{tian_scicode,
	address = {(Vancouver, Canada)},
	author = {Tian, Minyang and Gao, Luyu and Zhang, Shizhuo Dylan and others},
	booktitle = {Proc. Conf. on Neural Inf. Processing Systems},
	month = {Dec},
	title = {SciCode: A Research Coding Benchmark Curated by Scientists},
	year = {2024}}

@article{wilcoxon,
	author = {Frank Wilcoxon},
	issn = {00994987},
	journal = {Biometrics Bulletin},
	number = {6},
	pages = {80--83},
	publisher = {[International Biometric Society, Wiley]},
	title = {Individual Comparisons by Ranking Methods},
	url = {http://www.jstor.org/stable/3001968},
	volume = {1},
	year = {1945}}

@article{Gligoric2024In,
	author = {Gligori{\'c}, Kristina and Piccardi, Tiziano and Hofman, Jake M. and West, Robert},
	journal = {Harvard Data Science Review},
	number = {3},
	title = {In-{Class} {Data} {Analysis} {Replications}: Teaching {Students} {While} {Testing} {Science}},
	volume = {6},
	year = {2024}}

\end{document}